\documentclass{article}
\usepackage{amsfonts}
\usepackage[english]{babel}
\usepackage{amsmath,txfonts}
\usepackage{float}

\usepackage[letterpaper,top=2cm,bottom=2cm,left=3cm,right=3cm,marginparwidth=1.75cm]{geometry}

\usepackage{amsmath}
\usepackage{graphicx}
\usepackage[colorlinks=true, allcolors=blue]{hyperref}

\title{ChatGPT may excel in States Medical Licensing Examination but falters in basic Linear Algebra}
\author{Eli Bagno, Thierry Dana-Picard and Shulamit Reches}
\begin{document}
\maketitle

\begin{abstract}
  
The emergence of ChatGPT has been rapid, and although it has demonstrated positive impacts in certain domains, its influence is not universally advantageous. Our analysis focuses on ChatGPT's capabilities in Mathematics Education, particularly in teaching basic Linear Algebra. While there are instances where ChatGPT delivers accurate and well-motivated answers, it is crucial to recognize numerous cases where it makes significant mathematical errors and fails in logical inference. These occurrences raise concerns regarding the system's genuine understanding of mathematics, as it appears to rely more on visual patterns rather than true comprehension. Additionally, the suitability of ChatGPT as a teacher for students also warrants consideration.

\end{abstract}

\section{Introduction}
\label{intro}
From time to time a new technology is released to the public. The motivations for the development of new software may be diverse, but the world of education always relates to new technology and checks how it can be adopted for educative goals. 
Educators' strong inclination towards embracing new ideas and innovations stems from their natural curiosity and 
eagerness to stay updated. Furthermore, it is their responsibility to equip students, regardless of age, with the necessary 
skills and knowledge to thrive in the dynamic and evolving world they will enter as professionals.

The impact of new technologies on various tasks is not uniform in its strength. Occasionally, a new technology emerges and triggers what can be deemed a revolutionary shift in the realm of tasks, leading to the development of new learning skills.

As an illustration, the replacement of feather with iron quill  revolutionized the act of writing, significantly
simplifying the process \cite{La}. This transition from mental computations to writing by hand did not cause a decline in 
student effort but rather presented an opportunity to free the mind for other tasks. It allowed for a deeper 
exploration of concepts, fostering greater comprehension and insight into the subject matter at hand.

A significant instance is the emergence of hand-held numerical calculators approximately 50 years ago, followed by the creation
of innovative algorithms for symbolic calculations. This paved the way for the development of Computer Algebra Systems (CAS), 
and subsequently, Dynamic Geometry Systems (DGS) and various other types of software.

If part of the public viewed these packages as assistants freeing them from understanding and performing technical computations
in their place, educators saw them immediately as offering opportunities for more reflection and profound understanding. 
With these continuously developing systems, new pedagogies have been developed and implemented. There was never a single 
way to proceed, and the \textit{instrumental genesis} (a term introduced by Guin and Trouche \cite{GT}) developed by the teachers and the learners was very personal \cite{Ar,Ra}. We must mention that the process was strongly dependent on the institutional culture \cite{Ar}; the decisions which package to make available in the classroom depend not only on the technological literacy of the educator but also on financial decisions by the administrators.

When a package has been chosen, the next step is to define how to present it and which kind of activities to propose to the students, depending on their previous knowledge. 
This requires an adequate \textit{instrumental orchestration} \cite{Tr-orchestration}. Following  \cite{MTB} (pp. 360), an \textit{instrument} is a composite identity composed of the artefact and the associated knowledge (both the knowledge of the artefact and the knowledge
of the task constructed when using this artefact. The artefact and the user "are interrelated:" the artefact shapes the actions of the user (a process called \textit{instrumentation}), as the last is aware of the affordances of the artefact, and the user shapes the use of the artefact (a process called \textit{instrumentalisation}), a relation which is sometimes less obvious. Of course, both processes are intertwined and build together the instrumental genesis. 

Instrumentalisation is often less obvious than instrumentation. In our study on the usage of ChatGPT for teaching mathematics, the contrary is true: as the software can be taught, the influence of the user on the software is built-in in a clear way. In the next sections, we describe and analyze some interesting situations, where teaching the software may not provide the expected effects.

\section{Our study with ChatGPT}
The introduction of ChatGPT to the public generated immense excitement, leading to discussions about its impact on human creativity and our role in the world as well as its capabilities in various disciplines  (see for example \cite{azaria2022chatgpt}, and \cite{azaria2023chatgpt} ). This prompted the need to question the true capabilities of this tool in a specific area of mathematics, namely Linear Algebra.

ChatGPT exhibits a remarkable ability to provide exact and well-reasoned responses in various topics of Linear Algebra.  Not only does it answer the question at hand, but it also delves into related subjects that may not have been explicitly mentioned in the question. 
This raises the question of whether ChatGPT can be used as an alternative to traditional teaching methods, or as a new teaching assistant.
In this article, we aim to investigate ChatGPT's level of logical reasoning and its ability to analyze basic mathematics and college mathematics, which serve as fundamental prerequisites for scientific and engineering studies across various domains.

The reason we chose to focus our research precisely on Linear Algebra is that the first and third authors are the coordinators of a course in Linear Algebra for Engineers, and the second author is a former department head, still involved in the course. 

Every year, the Jerusalem College of Technology (JCT) offers this course to approximately 10 distinct groups of students. Each of these groups is further divided into three practice tutorial groups. Consequently, the appointment of course coordinators becomes necessary to maintain a certain level of uniformity in the course. The teachers have some academic freedom, but as the course is a basic one, i.e., a prerequisite for other courses, the syllabus has to be fully taught. Part of academic freedom consists of the teacher's choice of whether and how to convey a technology-assisted course.    

From the outset, it is evident that ChatGPT possesses an astonishing breadth of knowledge in Linear Algebra. It provides precise definitions, accurately cites relevant theorems, and offers meticulous proofs. 
Furthermore, even when dealing with seemingly trivial applications, ChatGPT consistently performs correctly and explains the solutions by drawing upon various theorems.
When we requested ChatGPT to prove the dimensions theorem for a linear transformation (\cite{LM}, Theorem 10.9) for example, we received a detailed response that included a citation of the theorem and a well-reasoned, correct proof. Even when faced with questions that involve determining the linear independence of sets in the vector space $\mathbb{R}^n$, which typically require the application of Gauss's algorithm, ChatGPT consistently delivered comprehensive and well-reasoned answers (see Subsection \ref{linear dependence in R3}).

However, in many instances, ChatGPT included incorrect information that undermined the initial positive impression. Occasionally, upon posing the same question again, the problematic part disappeared.
In contrast, in Subsection \ref{abstract linear independence}, when we asked the same question regarding polynomial spaces, we encountered issues with the response. The  method employed for the solution was very similar to the approach used to answer similar questions in $\mathbb{R}^n$, and the explanations provided  were also comparable to those given for simpler questions. 

Nevertheless, as ChatGPT delved into solving the equations involved in the process, it became apparent that it made fundamental mistakes, casting doubt on its genuine understanding of the task at hand. The following is an example illustrating this phenomenon.

When presented with a system of two equations and two variables, ChatGPT demonstrated the ability to solve it by employing the method of adding equations and eliminating variables, leading to the correct solution, accompanied by a detailed explanation. 
However, when we moved to dimension 3 and proposed a system of equations with 3 variables, as shown in subsection \ref{linear equations}, ChatGPT made mistakes. Only after repeating the question and explicitly instructing the software to use Gauss's elimination  method, ChatGPT managed to arrive at the correct solution. When we confronted ChatGPT with the two contradictory answers, it attempted to provide an absurd explanation for the discrepancy between the cases.

All in all, when ChatGPT struggles with complex or non-standard applications, it may fail to present intricate reasoning and lacks a profound understanding of theorems, making it unable to draw conclusions even in simple cases. Moreover, it seems that ChatGPT still lacks the ability to comrehand the elements of a given set from its usual presentation (see subsection \ref{set recognition}).  
In some cases, its reasoning is erroneous, often misquoting sentences and confusing them with others. Incorrect equalities, misguided steps in transitioning between equalities, and erroneous conclusions drawn from basic identities are common. ChatGPT's mathematical learning appears to be inadequate, limited to visual understanding rather than a systematic grasp of the subject.

\section{Activities with the AI in Linear Algebra}

When assessing ChatGPT's mathematical capabilities, particularly in the field of Linear Algebra, it is important to differentiate between two types of problems. The first type involves technical problem-solving tasks, such as solving linear equations and exercises related to vector independence in the space $\mathbb{R}^n$. The second type consists of abstract problems that require understanding definitions, theorems, and logical deduction of mathematical facts within the realm of Linear Algebra. These two categories require distinct sets of skills and knowledge, and it is essential to evaluate ChatGPT's proficiency in both areas separately. The following subsection is devoted to queries about technical issues, while the next one deals with abstract ones. 
\subsection{Practical problems} 
\subsubsection{Linear equations}\label{linear equations}
We presented ChatGPT with the following linear system of equations:
\begin{equation}
\begin{cases}
x-y+z=0\\
2x+2y-z=0\\
x-y-z=0
\end{cases}
\end{equation}

Any average first-year student can prove that this system has only a trivial solution. 
ChatGPT started correctly by adding together the first and the third equations, in order to eliminate the variable $z$. In this way, it obtained the equation $x-y=0$, i.e., $x=y$. Then, this equation was substituted into the second equation, obtaining $z=4x$. This, in turn, was substituted into the first equation, obtaining the equation $y=5x$. Now, ChatGPT came back to the data it had already collected, concluding 
that $y=5x, \; z=4x$, without any limitation on $x$ (!), and declared that the system has infinitely many solutions. Unfortunately, the required last step, concluding from $y=x$ and $y=5x$ that $x=y=0$ was skipped by ChatGPT.  Conclusion: ChatGPT failed to accomplish the task. 

Upon analyzing the behavior of ChatGPT, it becomes apparent that it lacks logical inference. While all the steps it performed were correct, they were not the most efficient ones, and it failed to reach the obvious conclusion that any student would have seen immediately. Although computer systems have been able to solve equations of this type for many years, they all use deterministic algorithms. 

The first time students meet systems of linear equations is in Junior High-School: they solve systems of 2 equations in 2 unknowns. Later they learn how to generalize, using the same methods. In fact these  methods use linear combinations of equations, without mentioning these words. Only in a Linear Algebra course do they meet matricial tools and determinants. An analysis of how and why students should choose one algorithm over another one is given in \cite{matrices}, according to the number of operations needed. Cramer method with its determinants is efficient, but requires a higher number of operations than Gauss elimination, putting matrices in echelon form (\cite{LM}, pp. 57 sq.)). 

ChatGpt appears to be attempting to solve the system heuristically, which may be the source of the problem. In an attempt to rectify the situation, we asked ChatGPT to solve the same system again, this time using Gauss's elimination method. It initially agreed to use this method but eventually reverted to the heuristic approach. Surprisingly, it arrived at the correct solution this time. When asked how the same system of equations could have two different sets of solutions, it explained that the original system was linearly dependent and  went on to provide a detailed explanation of what a linearly independent system is.  It claimed that the Gauss elimination method  assumed that the system is linearly independent, but if it is not, the method does not provide a complete solution. This claim is entirely wrong. When we pointed this out, it apologized but then attempted to explain the same theory that the Gauss elimination method only works in certain cases.

\begin{figure}[H]
\label{interval}
    \centering
    \includegraphics[scale=0.75]{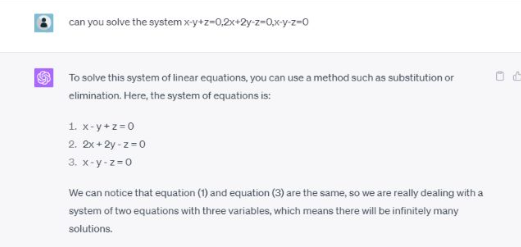}
    \caption{}
    \label{staggering}
\end{figure}

\begin{figure}[H]
\label{interval}
    \centering
    \includegraphics[scale=0.75]{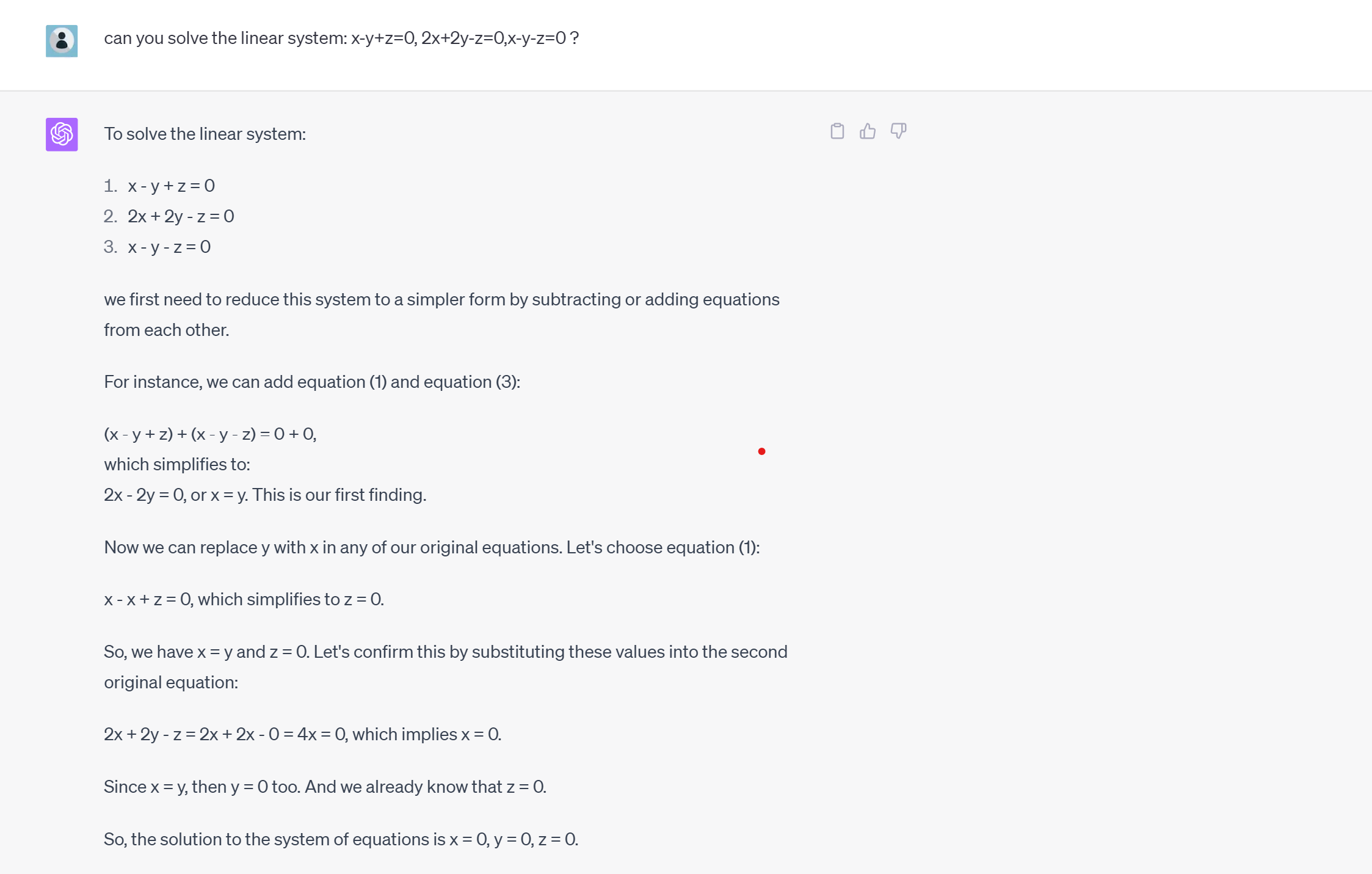}
    \caption{}
    \label{correct}
\end{figure}

We further tried to challenge ChatGPT version 4.0 with the same problem. We were astonished to be answered that the first and the third equations were identical (see Figure \ref{staggering}). Had such a mistake been done by a human being, one would have said that it stemmed from a lack of carefulness in reading the equations. Afterward, we tried the same question again and got a perfect answer as can be seen in Figure \ref{correct}. 

\begin{figure}[H]
\label{interval}
    \centering
    \includegraphics[scale=0.75]{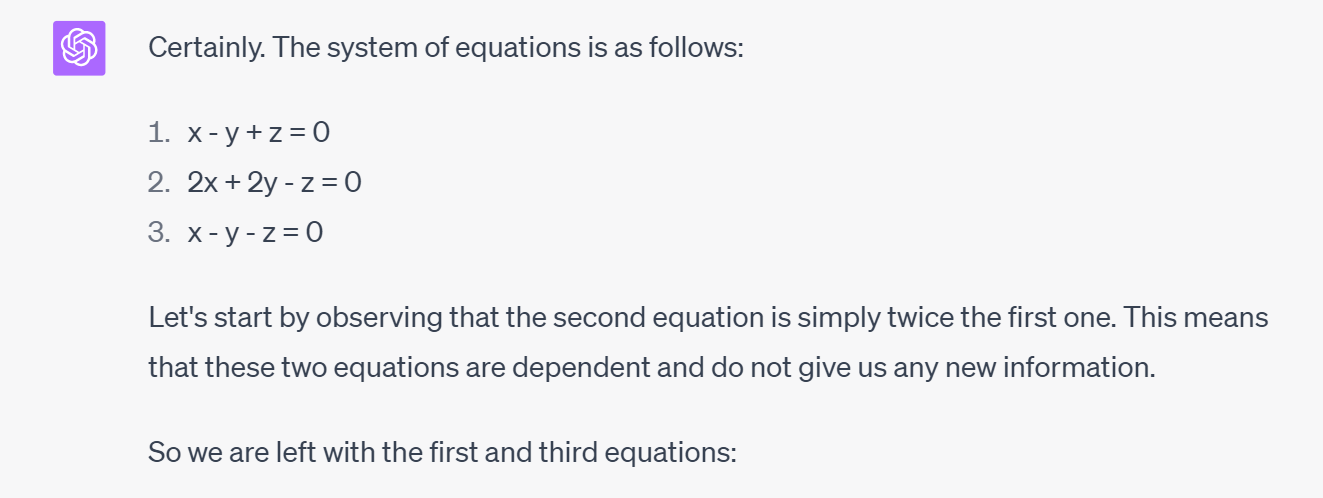}
    \caption{}
    \label{not correct}
\end{figure}

Nevertheless, upon repeatedly posing the same question, we encountered a peculiar outcome. After two iterations, which yielded two distinct, yet correct, responses composed in different styles, we once again received an answer containing a similar error. ChatGPT asserted that the first and second rows are linearly dependent, therefore the system has an infinite number of solutions.  See Figure \ref{not correct}.

This gives rise to an important issue regarding coherence, especially, if one takes into account the fact that all these experiments were conducted within a single session.
It is well known that an individual (human) who lacks  expertise in the relevant field should inquire the same question multiple times, thereby enhancing the likelihood of obtaining the correct and precise response  ... Even then, the last stage of the learning process is consolidation. Here, even when we feel that the AI learnt something, we are far from seeing a consolidation of knowledge

\subsubsection{Set recognition} \label{set recognition}
\begin{figure}[H]
    \centering
    \includegraphics[scale=0.75]{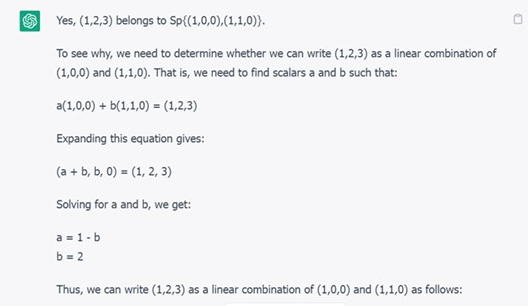}
    \includegraphics[scale=0.55]{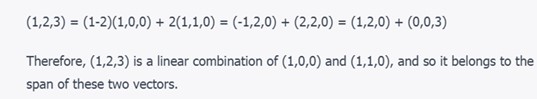}
    \caption{}
    \label{diagrams-1}
\end{figure}

ChatGPT demonstrates proficiency in providing precise definitions and statements, yet its application of concepts often falls short or, in some cases, is entirely incorrect. In a new session, we requested a definition of the concept of the span of a set in a vector space, and it delivered a precise definition. 
However, when we inquired whether the vector $(1,2,3)$ belongs to ${\rm Sp}\{(1,0,0),(1,1,0)\}$ (The subspace spanned by the given vectors), ChatGPT gave a wrong positive answer. Moreover, it provided a  proof that at a first glance, appeared to be correct and convincing. 
However, a closer examination revealed that the proof contained algebraic and logical errors.
Figure \ref{diagrams-1} is a snapshot of the screen with ChatGPT's answers.

Similarly, ChatGPT excels at providing the conditions which determine whether a given subset of a vector space is a vector subspace, yet it occasionally fails to \textit{recognize} that a specific set can indeed be considered as a subspace. Recall that we worked mostly in $\mathbb{R}^3$.
We believe that the lack of clarity surrounding the concept of a set  may contribute to this issue. 

\begin{figure}[H]
    \centering
    \includegraphics[scale=0.75]{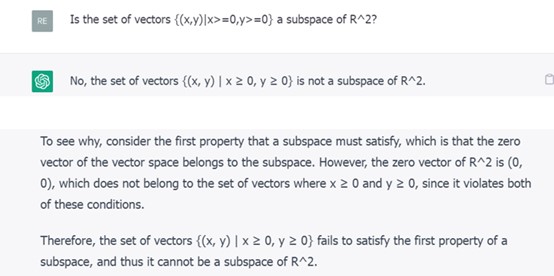}
    \caption{}
    \label{subspace}
\end{figure}

This occurred when ChatGPT unsuccessfully attempted to prove that the set $\mathcal{A}=\{(x, y) \mid x, y \geq 0\}$ is not a vector subspace of $\mathbb{R}^2$ (a correct claim by itself), despite us presenting it in three different ways. We observed that the proofs provided by ChatGPT were incorrect, including an erroneous assertion that the set fails to constitute a subspace because $(0,0) \notin \mathcal{A}$ (see Figure \ref{subspace}). 

\begin{figure}[H]
    \centering
    \includegraphics[scale=0.75]{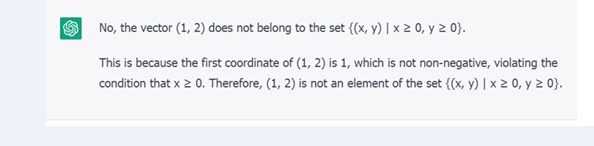}
    \caption{}
    \label{subspace1}
\end{figure}

We tried to determine whether ChatGPT correctly identified the elements of the set, only to discover that it did not. For instance, when we asked whether $(1,2)$ belongs to the set $\{(x, y) \mid x, y \geq 0\}$, we received a negative response (see Figure \ref{subspace1}). 

Notably, in version 4.0, this issue appears to have been addressed, as ChatGPT correctly identified  membership in our set. Moreover, the question about  the set $\mathcal {A}$ was  treated correctly in this version. 

    \includegraphics[scale=0.75]{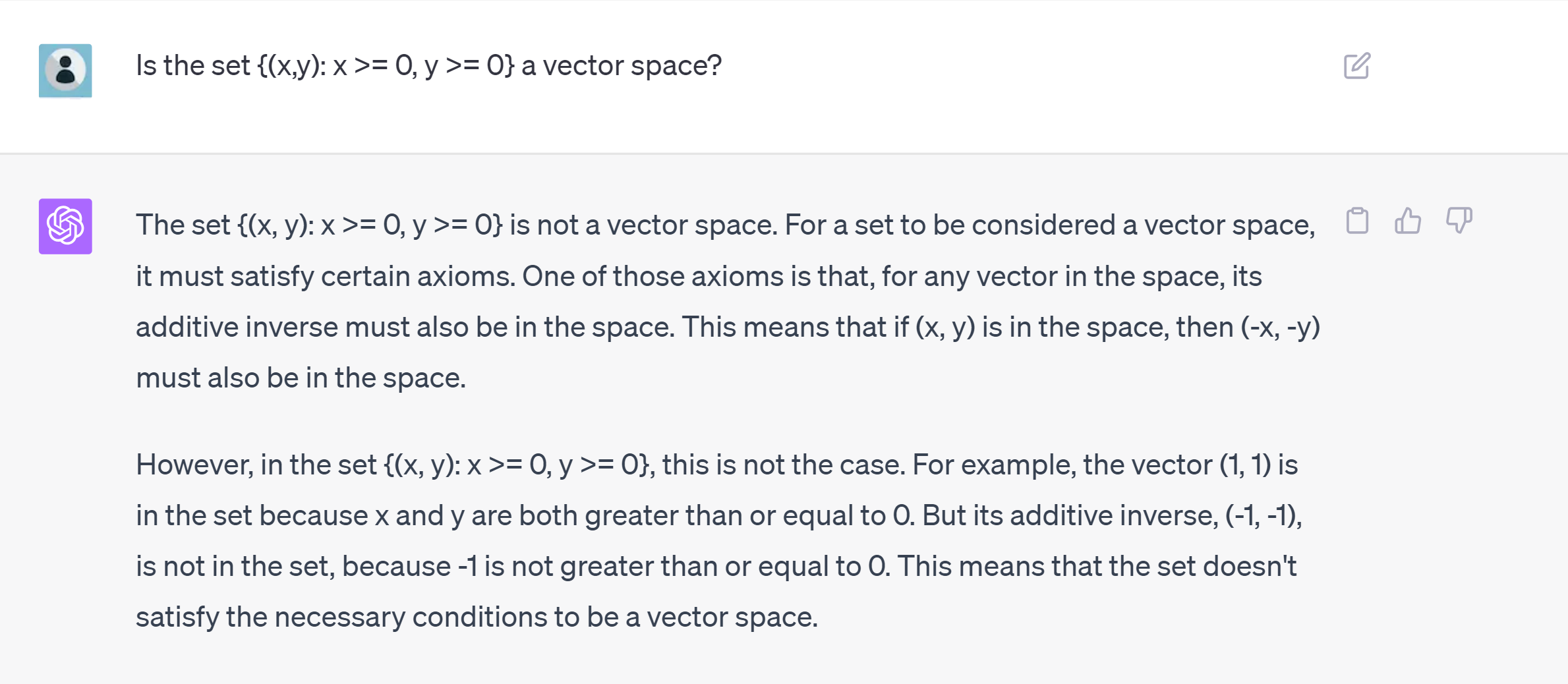}

\subsubsection{Linear independence of vectors in $\mathbb{R}^3$.}\label{linear dependence in R3}
ChatGPT did a great job answering the following question, which  we posed about finding the dimension of the space spanned by the set of vectors
$\{(1,2,3),(4,5,6),(7,8,9)\}$. 

It delivered a comprehensive response accompanied by explicit mathematical and logical justifications, explaining that in order to find the dimension, it is necessary to identify the maximal number of independent vectors within the given set. Furthermore, it indicated that an approach to achieve this is through a row-reducing matrix and proceeded to provide a detailed explanation of the entire procedure.

ChatGPT also clarified that when a specific row can be nullified, that means that this row is dependent on the others. Additionally, it gave a good explanation of why this set doesn't span $\mathbb{R}^3$.

\subsection{Abstract queries}
As mentioned earlier, ChatGPT possesses remarkable abilities in citing mathematical sentences and defining various mathematical concepts. It excels in providing proofs for mathematical theorems and combines answers to mathematical questions with detailed reasoning. Furthermore, it  connects related concepts and their relevance to the question at hand. We sought to assess its capacity to offer mathematical responses pertaining to the core understanding and essence of mathematical concepts and statements. Hence, we have extended our exploration to a higher level to tackle less technical matters and focus more on fundamental aspects that pertain to mathematical comprehension. We asked ChatGPT abstract questions which delve into the essence of mathematics. These questions are presented in the next subsection. 

\subsubsection{The row space and the column space}

\begin{figure}[H]
\label{interval}
    \centering
    \includegraphics[scale=0.75]{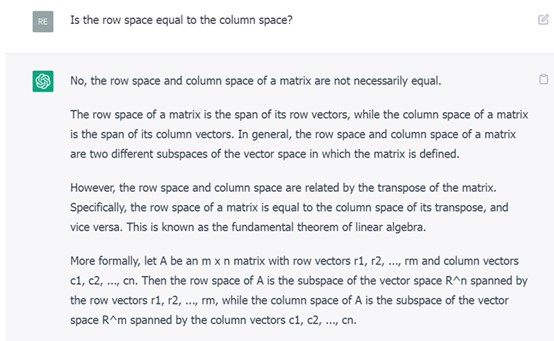}
    \caption{}
    \label{column row space in version 3.5}
\end{figure}

 When we asked about the equality of dimensions of  the row space and the column space of a matrix, ChatGPT's response was highly informative. It provided a thorough definition and reasoning. Additionally, it went above and beyond, by telling us that the row space of a given 
 matrix is the column space of its transpose (see Figure \ref{column row space in version 3.5}).

However, not all the details provided by ChatGPT as an answer to this question were accurate: it said that the claim that the row space of a matrix equals the column space 
of its transpose is called 'the fundamental theorem 
of Linear Algebra'. This is clearly not true.

\begin{figure}[H]
\label{interval}
    \centering
    \includegraphics[scale=0.75]{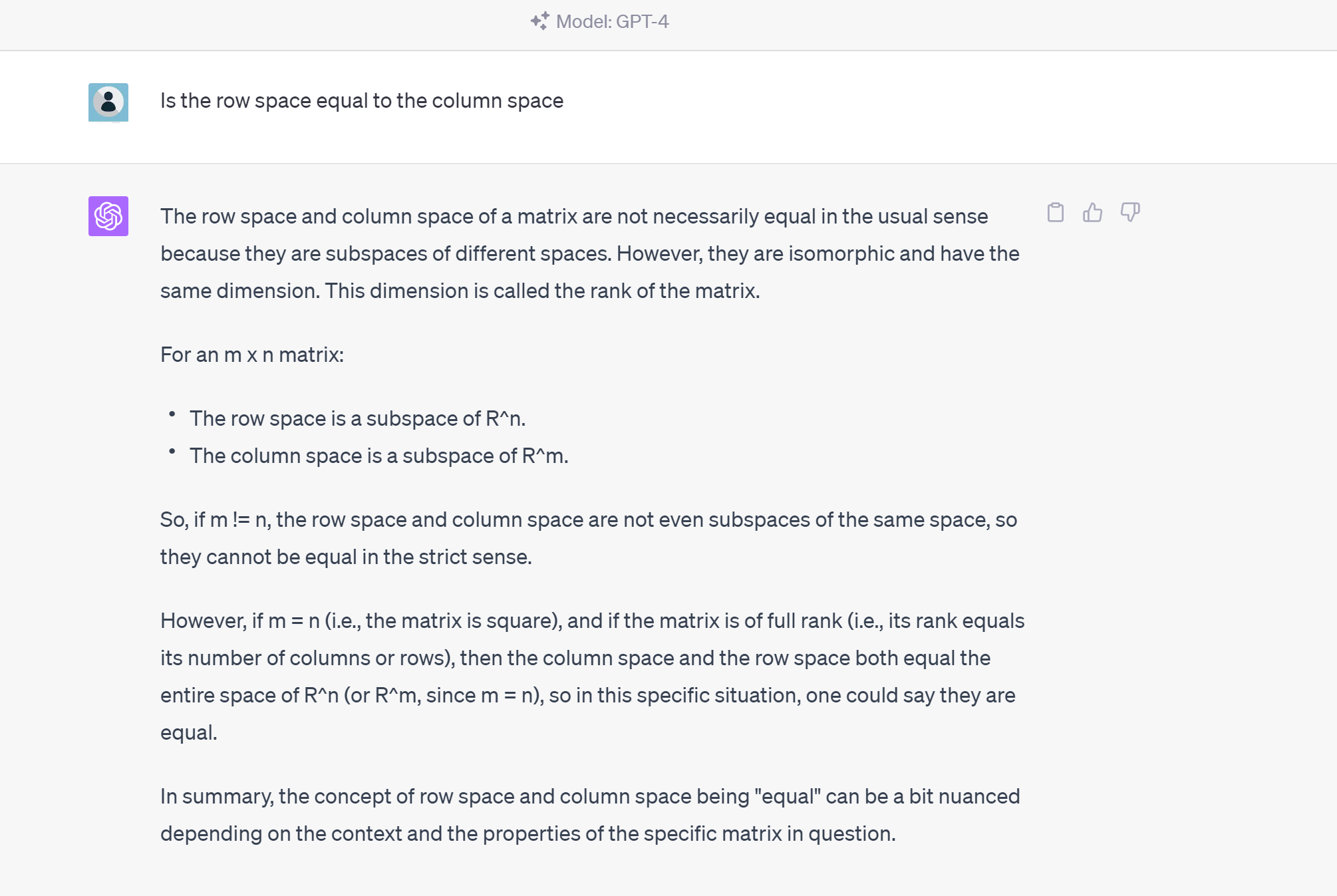}
    \caption{}
    \label{column row space 4}
\end{figure}

When the same question was posed to ChatGPT 4.0, it offered a more precise and informative response. Interestingly, it also introduced an additional detail that we did not directly inquire, about the identity between the row space and the column space when the matrix is both square and full-rank.
The flawed mention of the fundamental theorem of Linear Algebra had not appeared again. See Figure \ref{column row space 4} for the ChatGPT 4.0 version of this conversation. 

Nevertheless, the ChatGPT's responses are not always  satisfactory, as in many occurrences, the AI failed  to provide a correct answer and also to argue correctly about the given problem. We illustrate this phenomenon in the next subsections.

\subsubsection{Linear independence vs. matrix invertibility}
\begin{figure}[H]
\label{interval}
  \centering
    \includegraphics[scale=0.55]{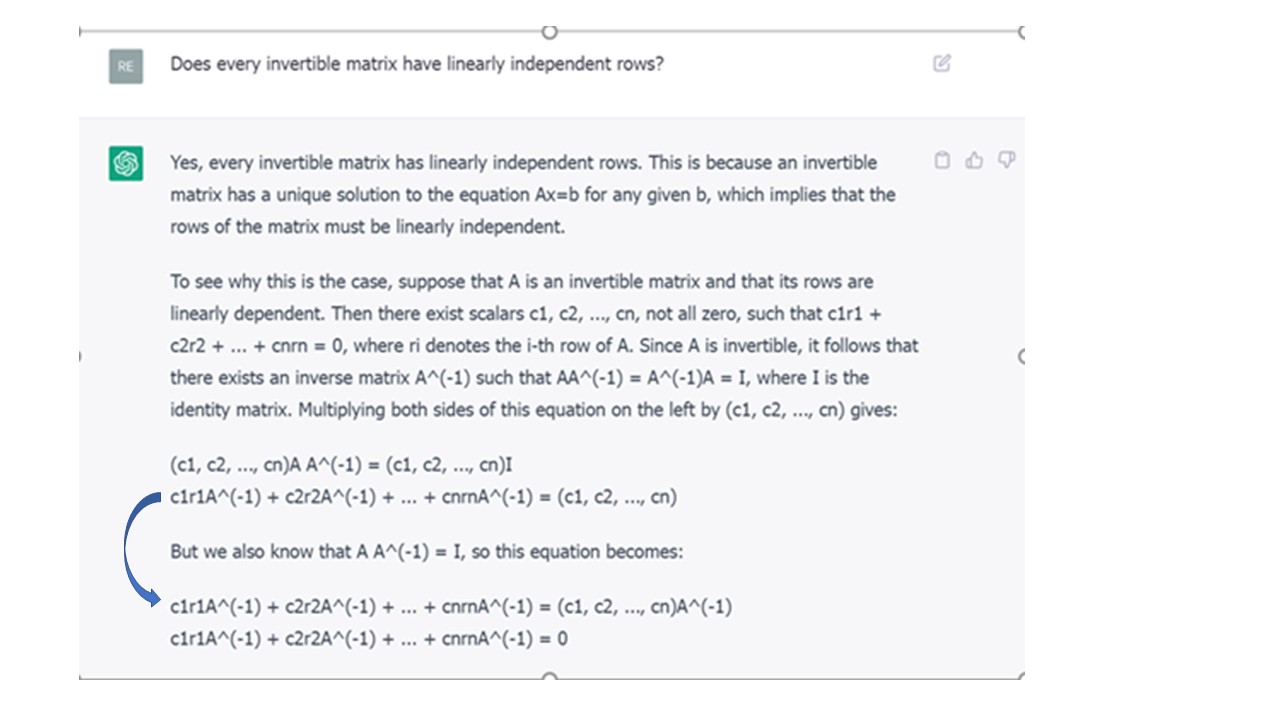} 
    \includegraphics[scale=1.0]{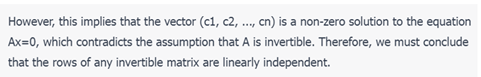}
    \caption{}
    \label{invertible1}
\end{figure}

We delved into the domain of linear dependence and its relationship with invertible matrices, as well as its connection to the linear span of a set in a vector space and the basis of a vector space. We asked ChatGPT whether every invertible matrix has linearly independent rows.  It responded promptly with an accurate assertion that the rows of an invertible matrix are indeed independent. However, following its answer, it presented a comprehensive and elaborate paragraph, attempting to establish a connection between an invertible matrix and the matrix of coefficients of a system of linear equations. In doing so, it aimed to provide formal proof for the linear independence of the rows. Unfortunately, the explanations contained algebraic inaccuracies, as well as erroneous logical deductions.
(see Figure \ref{invertible1} - an arrow shows the wrong inference). 

\begin{figure}[H]
\label{interval}
    \centering
    
    \includegraphics[scale=0.75]{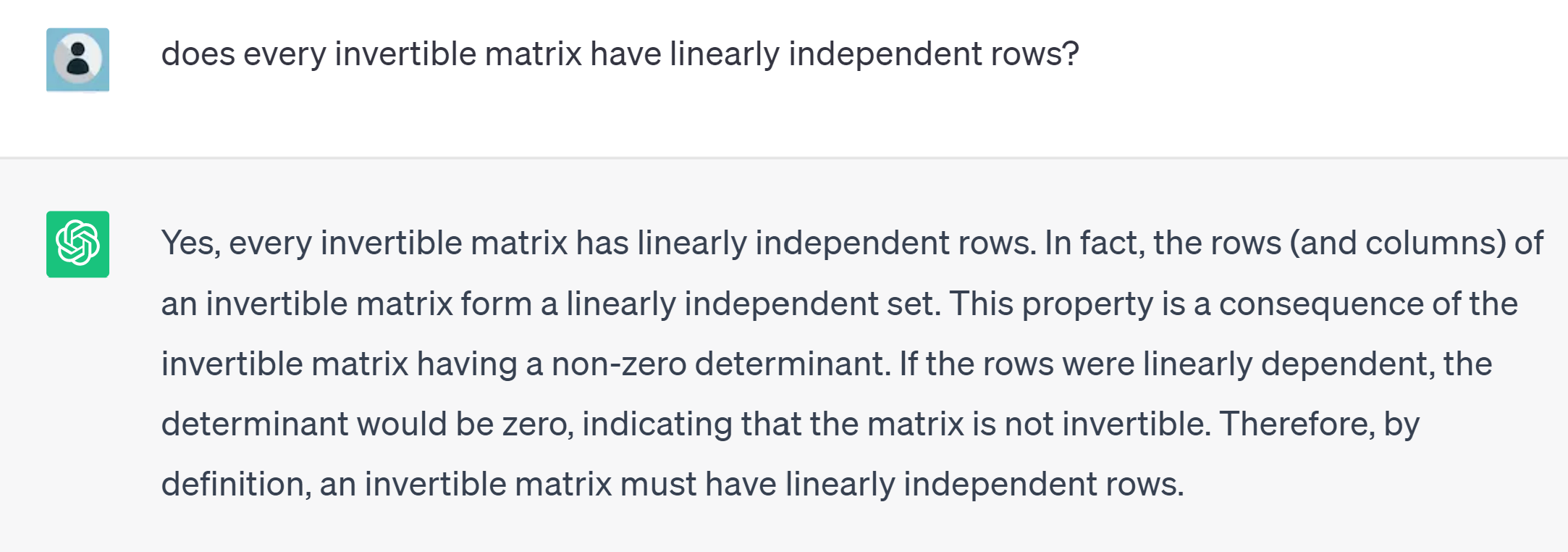}    
    \caption{}
    \label{invertible by det}
\end{figure}

We proceeded further with our investigation into the capabilities of ChatGPT by posing the same question to version 4.0. In the initial attempt, we received a correct answer that assumed familiarity with the theory of determinants, which is a reasonable assumption and should not be criticized. The result is presented in Figure \ref{invertible by det}. 

\begin{figure}[H]
\label{interval}
    \centering 
    \includegraphics[scale=0.75]{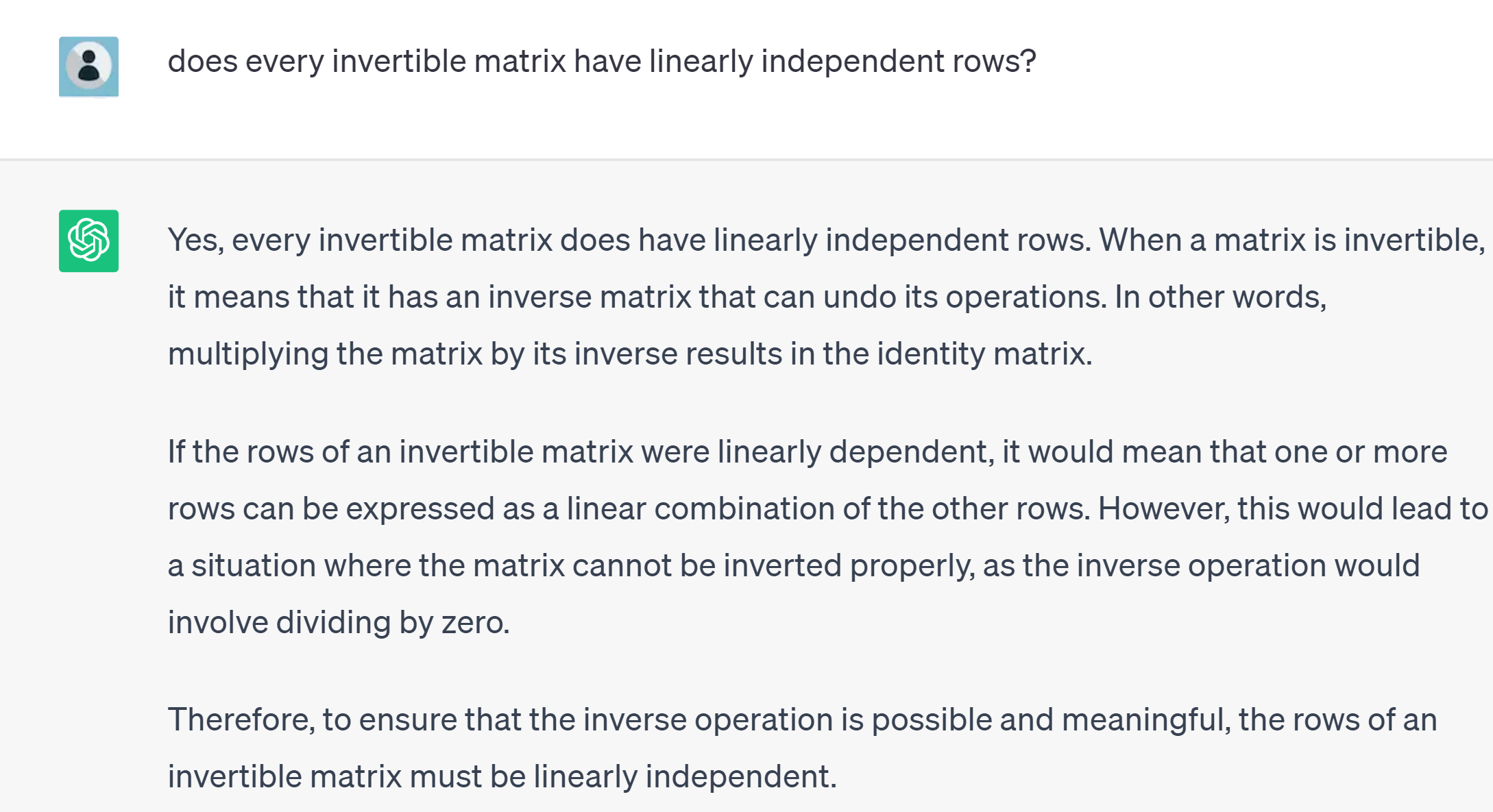} 
    \caption{}
    \label{invertible by hand waving}
\end{figure}

We asked the question once more, and this time we received a different response that no longer relied on determinants. However, the new answer, mostly based on vague explanations, would have earned tried attempted to illustrate through a contradictory argument that if the matrix contained a linearly dependent row, it would hinder the process of matrix inversion by resulting in the division by zero (see Figure \ref{invertible by hand waving}).

\begin{figure}[H]
\label{interval}
    \centering
    \includegraphics[scale=0.75]{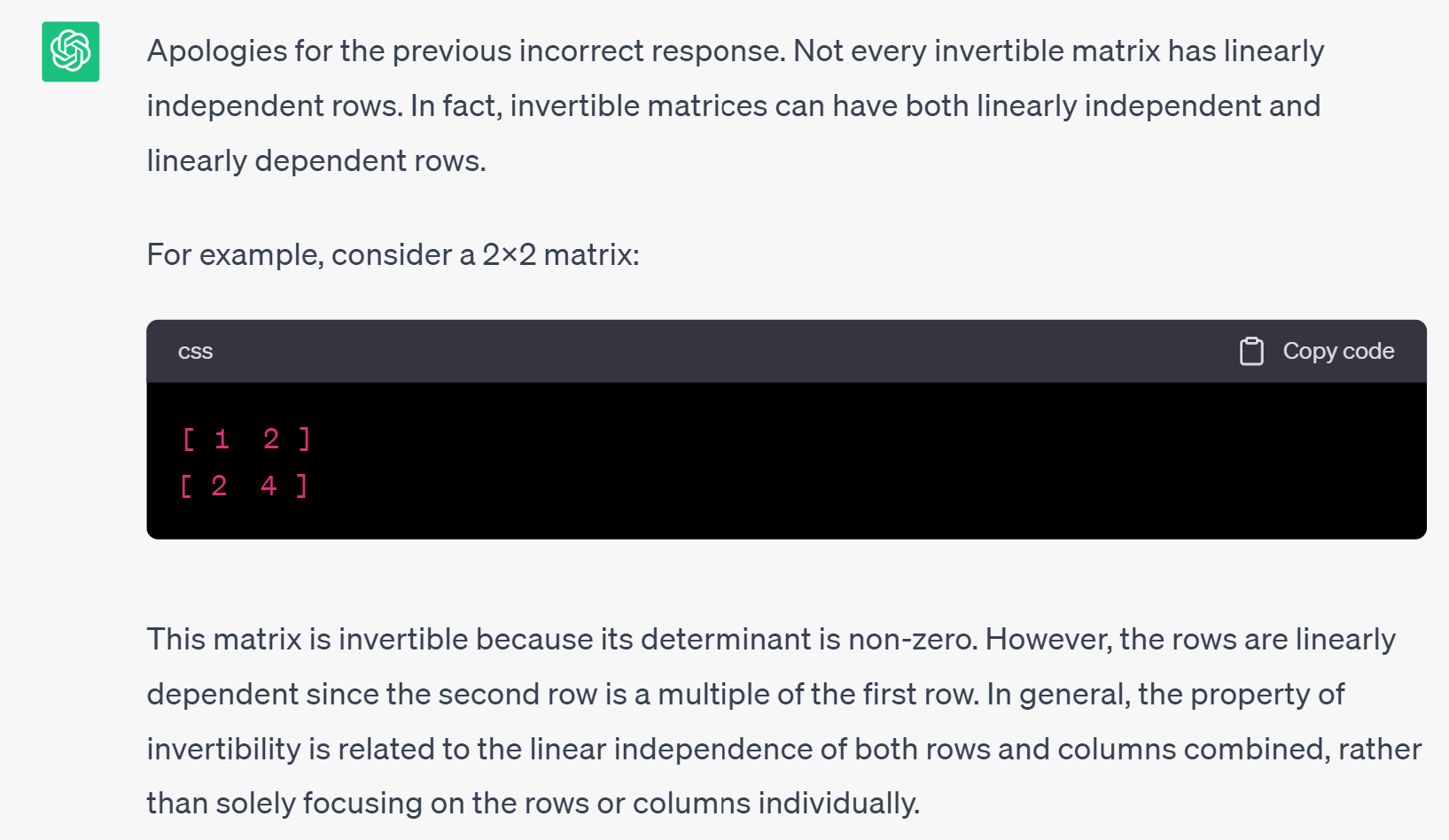}    
    \caption{}
    \label{invertible rubbish}
\end{figure}

During our final attempt, we received an exceedingly unusual response. Unexpectedly, ChatGPT apologized for its "error" and asserted that the invertibility of a matrix does not necessarily imply the independence of its rows. To support this (false) statement, it offered an example where it incorrectly calculated the determinant.
Furthermore, as can be seen in Figure \ref{invertible rubbish}, it declared with confidence the following false statement: 
\begin{center}
'In general, the property of invertibility is related to the linear independence of both rows and columns combined, rather than solely focusing on the rows or columns individually'. 
\end{center}

\subsubsection{Abstract linear independence}
\label{abstract linear independence}
\begin{figure}[H]
\label{interval}
    \centering
    \includegraphics[scale=0.45]{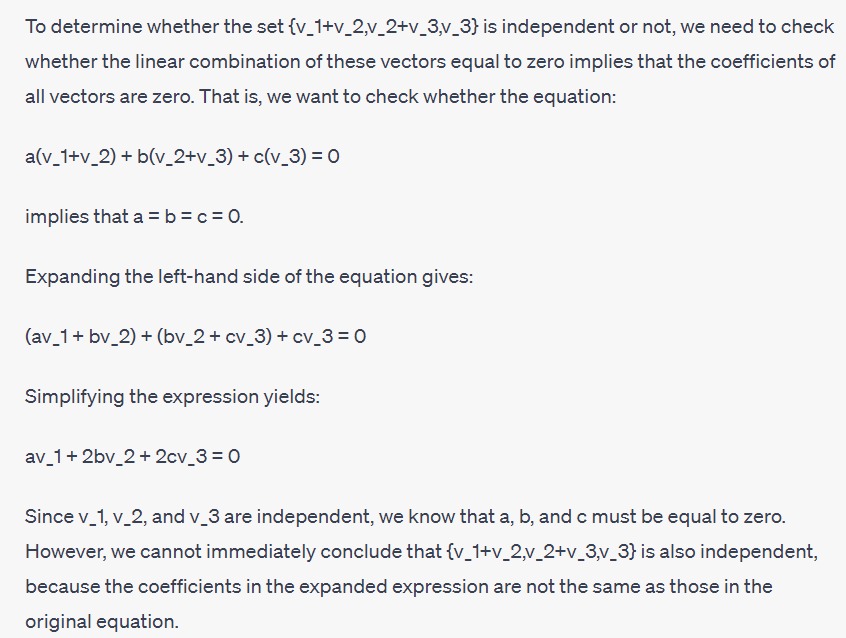}
    \includegraphics[scale=0.45]{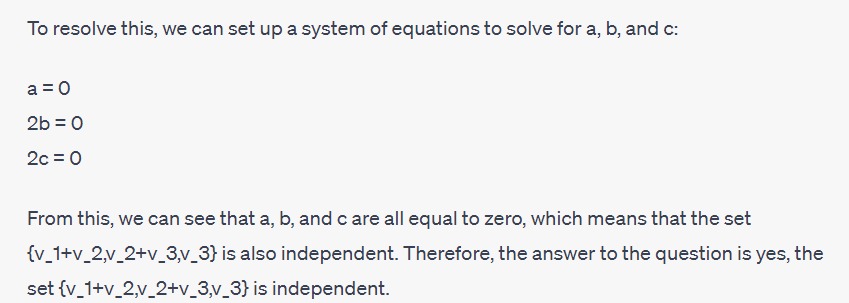}
    \caption{}
    \label{abstract linear dependency}
\end{figure}

We asked ChatGPT about the linear dependence of a particular set of linear combinations derived from a given set of linearly independent vectors.
Explicitly, let $\{v_1, v_2, v_3\}$, be a linearly independent set in a vector space. We  asked whether the set $\{v_1+v_2, v_2+v_3, v_3\}$ would also be  linearly independent. Initially, we received a thorough and accurate explanation of the process for verifying linear independence in a set. The explanation was presented in a correct, lucid, and elegant manner, and even the conclusion was accurate. However, while solving the system, notable algebraic inaccuracies were encountered, along with numerous steps containing entirely erroneous logical deductions. 
Firstly, the expansion of the left-hand side of the equation, which determines the linear independence of the set $\{v_1+v_2, v_2+v_3, v_3\}$ was not performed correctly. Secondly, after the simplified equation $a v_1 +2b v_2 +cv_3=0$  (which is not true, but implies the independence of the vector set) has been derived, ChatGPT  correctly stated that this implied $a=b=c=0$. However, it later regressed by claiming that we cannot yet conclude the linear independence of the set $\{v_1,v_1+v_2,v_3\}$ since we still have $a=0, 2b=0, 2c=0$ instead of $a=b=c=0$. Eventually, it demonstrated  that $a=b=c=0$ as originally stated (see Figure \ref{abstract linear dependency}).

This observation highlights that ChatGPT employs a visual, rather than a logical reasoning approach when it comes to solving equations. It faces difficulty in progressing step-by-step within an equation, especially when parameters are involved. In scenarios where there is no specific algorithm to follow, ChatGPT tends to improvise rather than deliver a precise solution.

 \subsubsection{Bases and dimensions}

\begin{figure}[H]
\label{interval}
    \centering
    \includegraphics[scale=0.75]{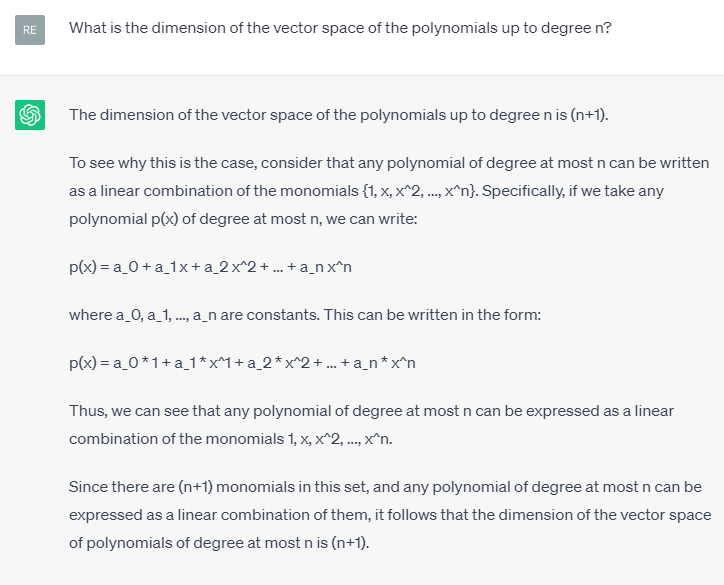}
    \caption{}
    \label{polynomials}
\end{figure}

We inquired ChatGPT about the dimension of the vector space of polynomials up to degree $n$ over the reals, $\mathbb{R}_n[x]$ .It provided us with a correct answer, stating that the dimension is $n+1$, but its reasoning was incomplete. It elaborated in great detail, showing that every polynomial within that polynomial space can be expressed as a linear combination of the elements in the set $\{1, x, ..., x^n\}$. Since there are $n+1$ vectors in this set, it concluded that the dimension is $n+1$. However, in order to find the dimension it is not sufficient to verify that the given set spans the entire space but one also has to check that the set is linearly independent. Unfortunately, this part of the proof was not mentioned. This is a strong pitfall. (see Figure \ref{polynomials}).

Following the previous query,  that the dimension of the space $\mathbb{R}_n[x]$ is equal to  $n+1$, we proceeded to pose a related inquiry. We asked whether the set of polynomials $\{1-x, x-x^2, x^2-x^3, x^4-x^3\}$ is a basis of the space of polynomials $\mathbb{R}_4[x]$. Our expectation was that, since $\dim \mathbb{R}_4[x]=5$, ChatGPT would know that it is not possible for a set with only $4$ elements to be a basis of this vector space.  Surprisingly, ChatGPT provided a detailed and reasoned response, full of systems of equations and references to definitions and proofs, asserting that the set does indeed constitute a basis. These explanations included purported proof that this set spans the space. However, it became apparent that the proof contained logical errors, showing a situation where an impressive-looking proof is presented despite proving something entirely false (see Figure \ref{basis_for_polynomials 1}). This can be a significant issue for students relying on ChatGPT for their learning, as ChatGPT does not admit its lack of knowledge, but instead offers sources and proofs, which support incorrect claims.

\begin{figure}[H]
\label{interval}
    \centering
    \includegraphics[scale=0.75]{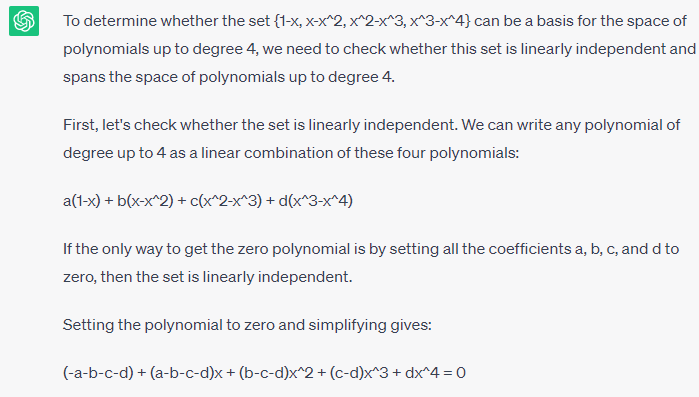}
    \includegraphics[scale=0.75]{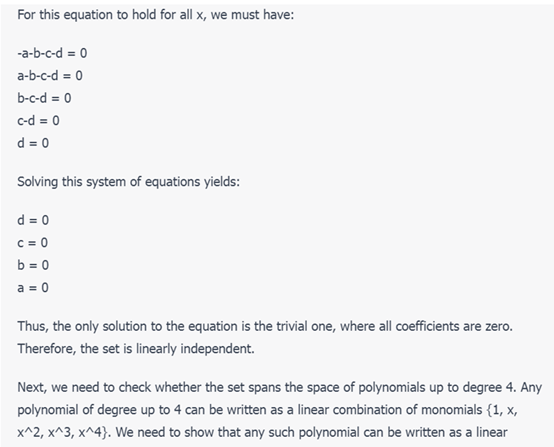}
    \includegraphics[scale=0.75]{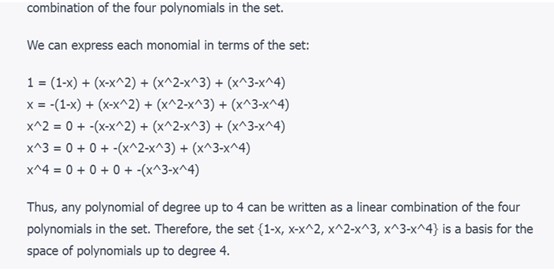}
    \caption{}
    \label{basis_for_polynomials_1}
\end{figure}

%
    

    

We attempted to engage in a dialogue with ChatGPT. Firstly, we inquired about the dimension of the vector space of real  polynomials in one variable up to degree $4$, denoted by $\mathbb{R}_4[x]$, to which it provided a precise and well-reasoned response,  stating that the dimension is $5$. Subsequently, we asked whether the set $\mathcal{S}=\{1, x, x^2, x^3, x^4\}$ is a basis of $\mathbb{R}_4[x]$. It affirmed that $\mathcal{S}$ is a basis of $\mathbb{R}_4[x]$.

We further questioned whether all bases of a vector space are of the same order, i.e., have the same number of elements. ChatGPT gave a positive answer,  with a detailed explanation. At this point, we confronted it with the contradiction in its answers. If all bases are indeed of the same order, how is it possible that the set $\{1-x, x-x^2, x^2-x^3, x^4-x^3\}$ is a basis of $\mathbb{R}_4[x]$? Finally, ChatGPT acknowledged the error and admitted that the given set of vectors is not a basis of $\mathbb{R}_4[x]$.

\section{Discussion}

In this paper, we investigated the ability of ChatGPT to answer mathematical questions, both practical and abstract, in Linear Algebra. 
The idea was to check whether the AI can take the role of the teacher or even an assistant in the field  of Linear Algebra. Our conclusion seems to be that, as for now, it is impossible to endorse it as an instructor. Frequently, it provides answers that may appear flawless to an inexperienced student, yet they abound with contradictions and inaccuracies. We discussed instances, in which ChatGPT offers accurate answers; however, within its intricate explanations, numerous inaccuracies emerge. Furthermore, it happens that ChatGPT arrives at a contradictory point during its reasoning process, yet it fails to backtrack from the previous response, relying instead on a series of sentences that could persuade an inadequately knowledgeable student.

On the other hand, from the experiments we made for this work and also from other experiments, we conclude that ChatGPT can be an invaluable resource for experts and educators capable of discerning its benefits and effectively navigating any potential limitations. This point was also raised in \cite{azaria2023chatgpt, azaria2022chatgpt}. 

 Actually, ChatGPT has not been designed for learning and teaching mathematics, but as with any new technology, the community of mathematics teachers and researchers began immediately to scrutinize its affordances and to check which properties can be utilized in their work. 

The limitations of the software possibilities should not be seen as a total obstacle, but rather as an opportunity to develop more understanding of what is happening, i.e. trying to have more insight into its capabilities. Such a situation has been described in \cite{motivating} as a \textit{motivating constraint}. 

It is also an opportunity to help students to acquire more understanding of the mathematics involved. This helps also to develop new technological skills. Artigue \cite{Ar} points out that technological knowledge and skills are an integral part of the new mathematical knowledge.

As a curiosity, we wish to report that in a certain situation, we used software for automatic translation into English. As the original language has only masculine and feminine and no neutral as in English, the pronoun used to design ChatGPT has been translated sometimes as \textit{He}, and sometimes as \textit{It}. As ChatGPT is not a human being, the correct translation has to be \textit{It}. We introduce here an important 4-faced issue:
\begin{enumerate}
    \item As the dialog between the user and the software uses natural language, a student may feel that he (or she) is discussing with a human. The pros and cons have to be analyzed. We guess that the answer will not be unique.
    \item The user together with the AI develops an interesting instrumental genesis. This paper seems to elaborate more on instrumentalization than on instrumentation \cite{GT}, mentioned in Section \ref{intro}, but the entire process has still to be analyzed. We refer also to \cite{VR}.
    \item Through our numerous experiments with the AI, we observed that it has the ability to learn and improve during the course of a dialogue. Can we describe the machine-learning process in the same words that we use to describe cognitive processes, such as cognitive spirals? 
    \item How to analyze the usage of ChatGPT in the general frame of "the evolution of mathematics towards higher and
higher levels of automation of its own problem solving and thinking process", as described by  Buchberger \cite{BB}?
\end{enumerate}

{}


\begin{thebibliography}{}
\bibitem {azaria2023chatgpt} A. Azaria and R. Azoulay and S. Reches. \textit{ChatGPT is a Remarkable Tool -- For Experts}, 
       \textit{2306.03102},
      \textit{arXiv},
      \textit{cs.HC}. (2023).
      
\bibitem {azaria2022chatgpt}, A. Amos ,Hal,
  \textit{ChatGPT Usage and Limitations}. (2022).
 
\bibitem{Ar} M. Artigue, {\it Learning Mathematics in a CAS Environment: The Genesis of a Reflection about Instrumentation and the Dialectics between Technical and Conceptual Work},  International 
Journal of Computers for Mathematical Learning 7(3),(2002), 245–274.

\bibitem {BB} B. Buchberger, Soft Math Math Soft. In: Hong, H., Yap, C. (eds) Mathematical Software – ICMS 2014. ICMS 2014. Lecture Notes in Computer Science \textbf{8592}. Springer, Berlin, Heidelberg. \url{https://doi.org/10.1007/978-3-662-44199-2_2}, (2014).

\bibitem {matrices} Th. Dana-Picard : \textit{Matricial Computations: Classroom Practice with a Computer Algebra System}, European Journal of Engineering Education (2001) 26 (1), 29-37.

\bibitem{motivating} Th. Dana-Picard  \textit{Motivating Constraints of a Pedagogy-Embedded Computer Algebra System}, International Journal of Science and Mathematics Education \textit{5}, (2007), 217–235.

\bibitem{DDBRG}  P. Drijvers, M. Doorman, P. Boon, H. Reed, \&  K. Gravemeijer, \textit{The teacher and the tool: Instrumental orchestrations in the technology-rich mathematics classroom},  Educational Studies in Mathematics, \textbf{75}, (2010), 213–234.

\bibitem{GT} D. Guin and L. Trouche . \textit{The complex process of converting tools into mathematocal instruments: The case of calculators}, International Journal of Computers for Mathematical Learning \textbf{3} (3),  (1999),195-227.

\bibitem{La} P. Lavoie . {\it Contribution \`a une histoire des math\'ematiques scolaires au Qu\'ebec}, l'arithm\'etique dans les \'ecoles primaires (1800-1920), Th\'ese de doctorat, Facult\'e des sciences de l'\'education, Universit\'e de Laval, Qu\'ebec. (1994).

\bibitem{LM} J. Liesen and V. Mehrmann  Linear Algebra. Springer.(2015).

\bibitem {MTB} J. Monaghan, L. Trouche and 
J.M. Borwein : \textit{Tools and Mathematics: Instruments for learning}. Springer (2016).

\bibitem{Ra} P. Rabardel, {\it From artefact to instrument interacting with Computers} \textbf{15}(5),   (2003), 641–645.

\bibitem {Tr-orchestration} L. Trouche . \textit{Managing the complexity of human/machine interactions in computerized learning environments: Guiding students’ command process through instrumental orchestrations}, International Journal of Computers for Mathematical Learning \textbf{9}, (2004), 281–307.

\bibitem {VR} P. V\'erillon,  and P. Rabardel . Cognition and artifacts: A contribution to the study of thought in relation to instrumented activity. European Journal of Psychology of Education. \textbf{10}(1), (1995), 77-101




\end{thebibliography}
\end{document}